\documentclass[runningheads]{llncs}
\usepackage[T1]{fontenc}
\usepackage{graphicx,verbatim}
\usepackage{soul}
\usepackage{multirow}
\usepackage{hyperref}
\usepackage{color}

\usepackage{microtype}
\usepackage{graphicx}
\usepackage{subcaption}
\usepackage{booktabs} 
\usepackage{multirow}
\usepackage{amsmath}
\usepackage{booktabs}
\usepackage{colortbl}
\usepackage{xcolor}
\usepackage{adjustbox}
\usepackage{pifont}
\definecolor{rowgray}{gray}{0.92}
\definecolor{winblue}{RGB}{235, 245, 255}
\definecolor{lightgray}{gray}{0.92} 
\usepackage{amsmath}
\usepackage{amssymb}
\usepackage{mathtools}
\usepackage{amsfonts}
\usepackage{dsfont}
\usepackage{bm}

\usepackage[most]{tcolorbox}
\definecolor{findingbg}{RGB}{255,242,217}

\newtcolorbox{findingbox}{
  enhanced,
  colback=findingbg,
  colframe=black,
  boxrule=1pt,
  arc=4mm,
  boxsep=0pt,
  left=10pt, right=10pt,
  top=7pt, bottom=7pt,
  width=\linewidth,
}

\newcommand{\finding}[2]{%
  \begin{findingbox}
    \textbf{\textit{Finding}~#1.}\enspace #2
  \end{findingbox}
}

\usepackage{tikz}
\usetikzlibrary{arrows.meta,positioning,shapes.geometric}

\begin{document}

\title{When Does RL Help Medical VLMs? Disentangling Vision, SFT, and RL Gains}
\titlerunning{MedBridgeRL}

\author{Ahmadreza Jeddi\inst{1,2,3*} \and
Kimia Shaban\inst{1,2,3*} \and
Negin Baghbanzadeh\inst{2,4*} \and
Natasha Sharan\inst{1} \and
Abhishek Moturu\inst{1,2,3} \and
Elham Dolatabadi\inst{2,4} \and
Babak Taati\inst{1,2,3}
}

\authorrunning{Jeddi et al.}

\institute{University of Toronto, Canada \and
Vector Institute, Canada \and
KITE Research Institute, University Health Network, Canada \and
York University, Canada \\
\email{ajeddi@cs.toronto.edu}
}

\renewcommand{\thefootnote}{}
\footnotetext{\textsuperscript{*} Equal Contribution}

  
\maketitle              

\definecolor{customred}{HTML}{ED028C} 
\begin{abstract}


Reinforcement learning (RL) is increasingly used to post-train medical Vision-Language Models (VLMs), yet it remains unclear whether RL improves medical visual reasoning or mainly sharpens behaviors already induced by supervised fine-tuning (SFT). We present a controlled study that disentangles these effects along three axes: vision, SFT, and RL. Using MedMNIST as a multi-modality testbed, we probe visual perception by benchmarking VLM vision towers against vision-only baselines, quantify reasoning support and sampling efficiency via Accuracy@1 versus Pass@K, and evaluate when RL closes the support gap and how gains transfer across modalities. We find that RL is most effective when the model already has non-trivial support (high Pass@K): it primarily sharpens the output distribution, improving Acc@1 and sampling efficiency, while SFT expands support and makes RL effective. Based on these findings, we propose a boundary-aware recipe and instantiate it by RL post-training an OctoMed-initialized model on a small, balanced subset of PMC multiple-choice VQA, achieving strong average performance across six medical VQA benchmarks. Project Page: \href{https://medbridgerl.github.io/}{\textcolor{customred}{https://medbridgerl.github.io}}



\end{abstract}

\section{Introduction}
Medical VLMs are emerging as a unified interface for clinical imaging. By jointly processing images and textual queries to generate accurate and interpretable responses, they enable applications in visual question answering, report assistance, and decision support. However, accuracy alone is insufficient for clinical deployment, which also demands reliability, transparency, and robustness across modalities and institutions. Motivated by recent progress in ``reasoning'' Large Language Models (LLMs), there has been increased interest in adapting post-training methods, particularly supervised fine-tuning (SFT) and reinforcement learning with verifiable rewards (RLVR), to encourage coherent reasoning traces and improve trustworthiness~\cite{chen2024towards,pan2025medvlm,wang2025medical,fan2025chestx,zhang2025patho,xia2025mmedagent}.

Most existing medical VLM-RL pipelines closely follow the broader RLVR literature: start from a general open-source VLM pre-trained on internet-scale multimodal data, then apply limited SFT and/or RL on comparatively small medical subsets due to the difficulty of acquiring high-quality verifiable rewards and curated reasoning data. While several methods report promising gains, the results are often uneven across modalities and tasks, and cross-modality generalization remains inconsistent, even for models trained on multiple imaging types~\cite{lai2026med,dai2025qoq,pan2025medvlm}. This leaves fundamental questions unanswered: how much of the observed improvement comes from the visual perception versus language-side alignment, how much is attributable to SFT rather than RL, and under what conditions RL is actually worth its cost in medical settings.

In parallel, recent studies in general LLM literature challenge the common assumption that RLVR inherently creates new reasoning capabilities beyond the base model. By probing Pass@K behavior, Yue et al.~\cite{yue2025does} argue that RL primarily reshapes the output distribution to sample correct solutions more efficiently, often without expanding the underlying support. Controlled experiments by Zhang et al.~ \cite{zhang2025interplay} further show that post-training gains depend critically on where the base model's competence boundary lies and how training data cover that boundary. In this work, we bring these questions to medical VLMs and aim to characterize: (i) perception limits across modalities, (ii) the \emph{reasoning-capacity boundary} via Pass@K, (iii) when RL improves sampling efficiency and whether these effects transfer across medical modalities.

\paragraph{\textbf{Contributions.}}
We use MedMNIST-v2~\cite{medmnistv2} as a controlled multi-modality testbed to disentangle the roles of vision, SFT, and RL in medical VLM post-training.  Specifically, we:

\begin{enumerate}
    \item Probe visual perception via vision-only linear evaluation to identify representation bottlenecks;
    \item Characterize each model's support boundary by comparing Accuracy@1 and Pass@K across modalities;
    \item Evaluate when GRPO-style RL helps medical VLMs by testing in-domain gains and within- and cross-modality transfer, under both base and SFT-initialized models; and
    \item Turn these findings into a staged recipe (bridge support first, then sharpen with RL) and validate it by post-training an OctoMed baseline on PMC-VQA~\cite{zhang2023pmc}, achieving strong results on multiple medical VQA benchmarks.
\end{enumerate}
\section{Related Work}

\paragraph{\textbf{Medical VLMs and post-training.}}
Medical VLMs have progressed through large-scale multimodal pre-training followed by domain adaptation, most commonly via SFT on medical image--text pairs, VQA instruction data, and report-style corpora~\cite{myronenko2025reasoning,ossowski2025octomed,chen2024towards,xu2025lingshu}. More recently, several works have adopted RLVR/GRPO-style post-training to encourage structured reasoning traces and improve answer reliability under limited supervision~\cite{pan2025medvlm,lai2026med,huang2025medvlthinker,zhang2025med,xu2025medground,liu2025beyond}. While many of these pipelines follow an \emph{SFT then RL} pattern, differences in data mixtures, reward designs, and modality coverage make it hard to isolate what drives improvements or to anticipate when RL will help.

\paragraph{\textbf{Does RL add reasoning beyond the base model?}}
Recent studies in the general LLM literature question whether RLVR expands reasoning capabilities or primarily improves sampling efficiency by reweighting solutions already present in the base model’s distribution~\cite{yue2025does,karan2025reasoning,zhang2025interplay}. We bring these questions to medical VLMs. Compared to standard \emph{SFT$\rightarrow$RL} pipelines, we (i) diagnose when RL can help using a boundary view (Accuracy@1 vs.\ Pass@K) alongside modality-wise perception probing, and (ii) propose a support-first recipe: raise Pass@K with targeted bridging only when needed, then apply RL as a sharpening stage rather than assuming RL is universally beneficial.

\section{Disentangling Vision, SFT, and RL in Medical VLMs}
\label{sec:method}

We study how SFT and RL post-training affect medical VLMs, and when these stages provide benefits beyond a strong base model. Our analysis is organized around three research questions, each addressed with controlled evaluations and summarized with a short finding. We first describe the shared experimental setting.

\paragraph{\textbf{Testbed.}}
We use MedMNIST-v2~\cite{medmnistv2}, a controlled suite covering three imaging modalities and twelve tasks at a unified input resolution (224$\times$224). For fair comparison, we evaluate on balanced subsets of the official test splits. MedMNIST is well-suited for our goals: it enables vision-only probing of representations, VLM-style MCQ evaluation derived from class labels, and efficient RLVR experiments due to its scale and standardized format.

\paragraph{\textbf{Models.}}
Our base model is Qwen2.5-VL-7B-Instruct~\cite{bai2025qwen2}, denoted as $M_{\text{Base}}$. To isolate the effect of medical SFT, we use OctoMed~\cite{ossowski2025octomed} as $M_{\text{SFT}}$. As a representative RL-posttrained medical VLM, we include QoQ-Med~\cite{dai2025qoq}, denoted as $M_{\text{RL}}$. Additional baselines are introduced in Sec.~\S\ref{sec:recipe_training} when evaluating our final training recipe and checkpoints.

\subsection{RQ1: How Strong Are the Visual Representations in Medical VLMs?}
\label{subsec:rq1}

Medical VLM performance can be limited by the vision tower itself. While prior work has probed contrastive encoders (e.g., CLIP~\cite{radford2021learning}) for medical transfer~\cite{baghbanzadeh2025advancing}, similar analysis is less common for end-to-end autoregressive VLMs such as Qwen. We therefore first measure whether the underlying visual representations are already separable for medical tasks.

We freeze each model's vision encoder (ViT) and evaluate representation quality with linear probing on MedMNIST-v2. We also include MedViT-v2~\cite{manzari2025medical} as a strong vision-only reference.

\autoref{tab:vit_only_results} reports probe accuracy for $M_{\text{Base}}$, $M_{\text{SFT}}$, and $M_{\text{RL}}$ across modalities. We use these scores to diagnose whether VLM failures are primarily perception-limited: when a model’s frozen ViT cannot linearly separate classes for a dataset, gains from SFT or RL are naturally bounded, whereas strong probe accuracy suggests errors stem from language-side alignment and decoding/sampling rather than visual features.

\finding{1}{The base model already has reasonably separable visual features on many MedMNIST tasks, and medical SFT improves them further, especially on weaker datasets. RL does not consistently improve ViT probe accuracy, suggesting its effects are mainly on sampling/alignment rather than visual representations. Several datasets remain far below the MedViT-v2 reference, indicating perception bottlenecks that cap downstream gains.}

\begin{table*}[t]
\centering
\small
\setlength{\tabcolsep}{4pt}
\caption{
Linear probing on frozen vision encoders across MedMNIST-v2 tasks, grouped by modality. MedViT-v2 is included as a strong vision-only reference.
}
\resizebox{\textwidth}{!}{%
\begin{tabular}{l|cc cc cc cc cc cc}
\toprule
\multirow{2}{*}{\textbf{Method}}
& \multicolumn{6}{c}{\textbf{Radiology}}
& \multicolumn{3}{c}{\textbf{Microscopy}}
& \multicolumn{3}{c}{\textbf{Visible Light Photography}} \\
\cmidrule(lr){2-7}
\cmidrule(lr){8-10}
\cmidrule(lr){11-13}
& Breast & Chest & Pneumonia & OrganA & OrganC & OrganS
& Blood & Path & Tissue
& Derma & OCT & Retina \\
\midrule
\rowcolor{winblue}
MedViT-V2~\cite{manzari2025medical}
& 91.00 & 96.70 & 97.30 & 97.30 & 96.10 & 85.10
& 98.70 & 97.10 & 71.60
& 81.70 & 95.20 & 57.80 \\
\midrule
M\textsubscript{Base}
& 80.95 & 13.33 & 92.52 & 85.27 & 79.31 & 65.70
& 96.19 & 89.12 & 48.26
& 65.22 & 84.30 & 44.00 \\
\midrule
M\textsubscript{SFT}
& 84.52 & 16.51 & 93.35 & 89.01 & 83.59 & 70.90
& 96.19 & 93.35 & 51.99
& 67.08 & 91.10 & 52.00 \\
\midrule
M\textsubscript{RL}
& 75.00 & 16.19 & 92.95 & 85.52 & 80.35 & 64.96
& 96.86 & 88.50 & 52.88
& 69.57 & 84.80 & 51.00 \\
\bottomrule
\end{tabular}
}
\label{tab:vit_only_results}
\end{table*}

\subsection{RQ2: What Is the Reasoning-Capacity of Medical VLMs?}
\label{subsec:rq2}

Even when visual representations are adequate (RQ1), medical VLMs can underperform because correct answers exist in the model’s distribution but are not produced reliably under greedy decoding. We capture this as a support boundary by comparing single-sample accuracy to multi-sample success, and ask whether post-training increases a model’s support or mainly improves sampling efficiency.

For each MedMNIST task with $C$ classes, we convert classification into a multiple-choice VLM prompt with $C$ options and require the model to output the option letter. We report (i) Accuracy@1 under greedy decoding and (ii) Pass@K, the probability that at least one of $K$ independent samples is correct. We use $K \in \{1,2,4,8,16\}$ with temperature $0.7$ and top-$p$ $0.9$. Outputs are constrained to \texttt{<think>} and \texttt{<answer>} tags;  we verify the correctness of the predicted choice using Qwen2.5-VL-32B-Instruct.

\begin{figure*}[t]
    \centering
    \resizebox{\textwidth}{!}{%
        \includegraphics{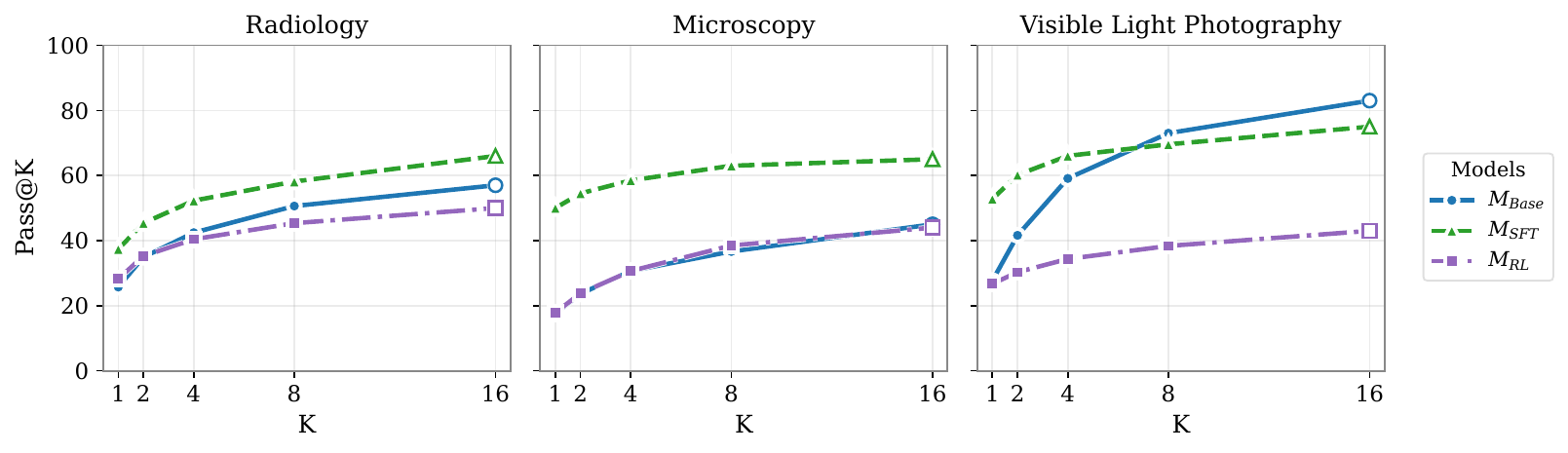}
    }
    \caption{Pass@K curves on MedMNIST-v2, grouped by modality.}
    \label{fig:pass_k_vlms}
\end{figure*}

\autoref{tab:vlm_results} reports Accuracy@1 across tasks, and \autoref{fig:pass_k_vlms} shows Pass@K curves grouped by modality. Together, they reveal how much latent support exists beyond greedy decoding and how different post-training stages shift the boundary.

\begin{table*}[t]
\centering
\small
\setlength{\tabcolsep}{4pt}
\caption{
VLM evaluation with Accuracy@1 under greedy decoding on MedMNIST-v2 tasks, grouped by modality.
}
\resizebox{\textwidth}{!}{%
\begin{tabular}{l|cc cc cc cc cc cc}
\toprule
\multirow{2}{*}{\textbf{Method}}
& \multicolumn{6}{c}{\textbf{Radiology}}
& \multicolumn{3}{c}{\textbf{Microscopy}}
& \multicolumn{3}{c}{\textbf{Visible Light Photography}} \\
\cmidrule(lr){2-7}
\cmidrule(lr){8-10}
\cmidrule(lr){11-13}
& Breast & Chest & Pneumonia & OrganA & OrganC & OrganS
& Blood & Path & Tissue
& Derma & OCT & Retina \\
\midrule
M\textsubscript{Base}
& 54.99 & 8.70 & 50.53 & 14.39 & 10.94 & 13.14
& 13.16 & 26.48 & 10.39
& 33.54 & 32.62 & 20.19 \\
\midrule
M\textsubscript{SFT}
& 79.17 & 13.22 & 86.00 & 18.66 & 16.03 & 19.38
& 76.00 & 64.53 & 12.66
& 52.33 & 82.78 & 22.19 \\
\midrule
M\textsubscript{RL}
& 69.79 & 8.03 & 62.41 & 10.91 & 8.94 & 10.75
& 13.95 & 34.09 & 8.96
& 27.48 & 28.72 & 20.62 \\
\bottomrule
\end{tabular}
}
\label{tab:vlm_results}
\end{table*}

\finding{2}{Across tasks, Accuracy@1 is often far below Pass@K, indicating substantial latent support that greedy decoding fails to realize. Medical SFT raises both Accuracy@1 and Pass@K, consistent with improved coverage and alignment. In contrast, the RL-post-trained baseline does not consistently improve Accuracy@1 on MedMNIST and frequently reduces Pass@K, suggesting distributional sharpening without expanding support and, in some cases, a narrower boundary.}

\begin{figure*}[t]
    \centering
    \resizebox{\textwidth}{!}{%
        \includegraphics{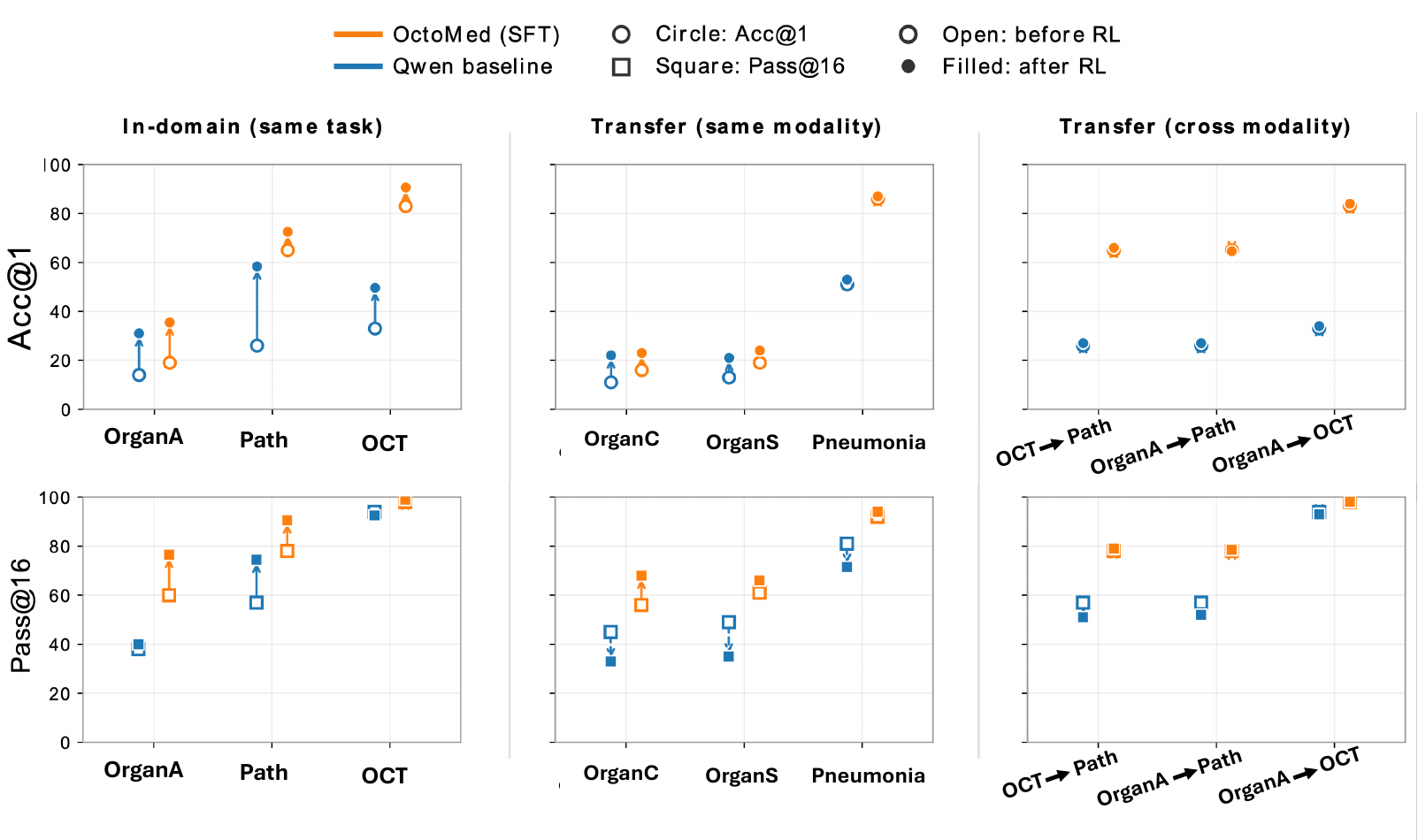}
    }
    \caption{Before/after RL changes in Acc@1 and Pass@16 from $M_{\text{Base}}$ and $M_{\text{SFT}}$ across in-domain, within-modality, and cross-modality evaluations.}
    \label{fig:rl_gains}
\end{figure*}

\subsection{RQ3: When Does RL Help Medical VLMs?}
\label{subsec:rq3}

RQ1--RQ2 suggest a simple picture. The base VLM has seen many of these modalities during large-scale pretraining, but coverage is uneven. A large medical SFT stage (OctoMed) improves visual perception and raises Pass@K across tasks. In contrast, an off-the-shelf RL medical checkpoint (QoQ-Med) does not reliably transfer to MedMNIST and can reduce Pass@K, which is consistent with the idea that RL often reshapes the distribution rather than adding new capability~\cite{yue2025does}. Here we ask when RL is actually helpful in a controlled setting, and what it does to both accuracy and support.

\paragraph{\textbf{Setup.}}
We run targeted RL post-training on one task per modality using the training split: OrganAMNIST (radiology), PathMNIST (microscopy), and OCTMNIST (visible-light photography). For each task, we train two RL variants initialized from (i) the base model $M_{\text{Base}}$ and (ii) the SFT-bridged model $M_{\text{SFT}}$. We use a consistency-aware GRPO pipeline~\cite{chen2025grpo,jeddi2025puzzle} instead of vanilla GRPO to stabilize optimization on small medical data. We evaluate three regimes: (1) in-domain (same task as RL training); (2) within-modality transfer by training on OrganAMNIST and testing on OrganCMNIST/OrganSMNIST (same family, different views) and PneumoniaMNIST (larger shift); and (3) cross-modality transfer with three shifts: OCT$\rightarrow$Path, OrganA$\rightarrow$Path, and OrganA$\rightarrow$OCT. Throughout, we report Acc@1 and Pass@16 before and after RL.

\paragraph{\textbf{Results.}}
\autoref{fig:rl_gains} shows that RL mainly acts as sharpening: it improves Acc@1 by placing more mass on answers that already exist in the model’s support. Consistent with \cite{zhang2025interplay}, even a modest bridge is enough to produce clear in-domain gains, but the size of the gain is bounded by the starting Pass@16. For small within-modality shifts (OrganA$\rightarrow$OrganC/OrganS), RL often improves Acc@1, while base-initialized RL more frequently trades off Pass@16. For the larger within-modality shift (OrganA$\rightarrow$Pneumonia), gains are smaller and Pass@16 drops become more common for $M_{\text{Base}}$. Under cross-modality shifts, RL produces little change in Acc@1 and can slightly reduce Pass@16, especially when starting from the unbridged base model.

\finding{3}{RL is most effective when the model already has non-trivial support: it sharpens the distribution, improves Acc@1 (sampling efficiency), and narrows the Acc@1--Pass@K gap in-domain and for small within-modality shifts. When support is weak (large or cross-modality shifts), accuracy gains are limited and Pass@K can drop, especially from an unbridged base model.}

\begin{figure*}[t]
    \centering
    \resizebox{\textwidth}{!}{%
        \includegraphics{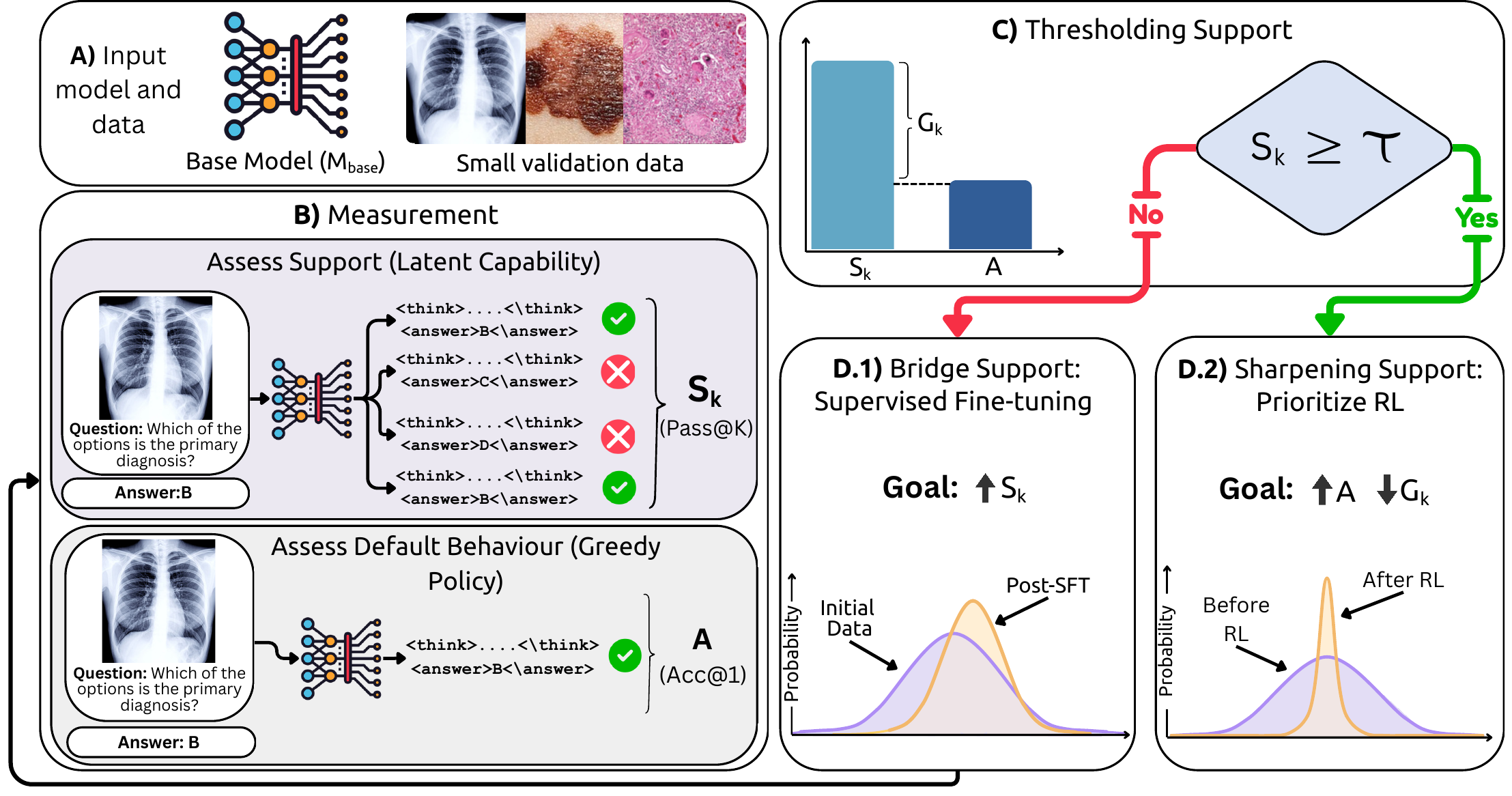}%
    }
    \caption{Overview of our boundary-aware recipe. We first diagnose support using Pass@K and Acc@1 then decide between bridging versus RL sharpening.}
    \label{fig:method_diagram}
\end{figure*}

\section{From Analysis to Practice: A Recipe for RL Post-Training}
\label{sec:recipe_training}

The previous section suggests that RL is most useful once a model has non-trivial support on the target distribution. In practice, RL can efficiently improve Acc@1 by sharpening the model's output distribution, but it remains constrained by the model's underlying Pass@K. We now turn these observations into a simple decision rule and apply it to real medical VQA benchmarks (Fig.~\ref{fig:method_diagram}).

\paragraph{\textbf{Support and sharpening.}}
For a task distribution $\mathcal{D}$ and a fixed sampling budget $K$, we define:
\begin{equation}
S_K(\mathcal{D}) = \mathrm{Pass@K}(\mathcal{D}), \qquad 
A(\mathcal{D}) = \mathrm{Acc@1}(\mathcal{D}), \qquad
G_K(\mathcal{D}) = S_K(\mathcal{D}) - A(\mathcal{D}),
\end{equation}
where $S_K$ measures latent support, $A$ measures the default behavior under greedy decoding, and $G_K$ is the support gap.

\paragraph{\textbf{Recipe (boundary-aware post-training).}}
Given a target reliability threshold $\tau$ on a task distribution $\mathcal{D}$, we use the following staged rule:
\begin{enumerate}
    \item \textbf{Diagnose support.} Estimate $S_K(\mathcal{D})$ and $A(\mathcal{D})$ on a small validation set.
    \item \textbf{Bridge if support is weak.} If $S_K(\mathcal{D}) < \tau$, prioritize bridging: add task- or modality-proximal data and perform SFT (or continued training) to raise $S_K$. This step aims to expand coverage rather than to sharpen.
    \item \textbf{Sharpen once support is sufficient.} If $S_K(\mathcal{D}) \ge \tau$, apply RL  to improve sampling efficiency, increasing $A(\mathcal{D})$ while monitoring $S_K$ and $G_K$ to avoid over-sharpening that collapses support.
\end{enumerate}
This formulation matches the behavior observed in RQ3: bridging increases support, while RL mainly converts support into higher Acc@1.

\paragraph{\textbf{Instantiation.}}
We instantiate our recipe on top of OctoMed-7B, which already provides a strong starting point via large-scale medical SFT. We then perform RL post-training on a small, balanced set of 8{,}000 multiple-choice questions sampled from the training split of PMC-VQA~\cite{zhang2023pmc}. To balance modalities, we first use Qwen2.5-VL-32B-Instruct to classify each question into common clinically established imaging categories (e.g., MRI, CT, ultrasound, X-ray, microscopy, dermatology, OCT), and assign a \texttt{none} label when the modality is uncommon, ambiguous, or not clearly specified. We then construct a balanced modality-balanced subset based on these predictions. Finally, we apply GRPO-style RL with a consistency-aware variant to stabilize training under limited data.

\paragraph{\textbf{Evaluation on medical VQA.}}
We evaluate the resulting RL post-trained model on six multimodal medical benchmarks, and compare against strong public baselines (Table~\ref{tab:med_vqa}). The results follow the expected pattern from our analysis: starting from an already-bridged checkpoint enables RL to improve accuracy without the failure modes observed when RL is applied to weak-support bases. Overall, our model achieves the strongest average performance among the listed Qwen2.5-VL-based medical VLM baselines under the same decoding setting.

\begin{table*}[t]
\centering
\small
\setlength{\tabcolsep}{4pt}
\caption{
Performance of medical VLMs on multimodal medical benchmarks. Best results are in bold; second-best are underlined.
}

\resizebox{\textwidth}{!}{%
\begin{tabular}{l|ccc | ccc | c}
\toprule

\textbf{Method} & PMC~\cite{zhang2023pmc} & MMMU~\cite{yue2024mmmu} & MedX-M~\cite{zuo2025medxpertqa}
& PathVQA~\cite{he2020pathvqa} & SLAKE~\cite{liu2021slake} & VQA-Rad~\cite{lau2018dataset} & Avg. \\

\midrule

Qwen2.5-VL-7B-Instruct~\cite{bai2025qwen2}
& 53.00 & 55.29 & 23.50 & 65.00 & 68.00 & 74.00 & 56.47 \\
\midrule

QoQ-Med-7B~\cite{dai2025qoq}
& 51.00 & \underline{61.18} & 25.00 & 63.50 & 68.50 & 74.00 & 57.20 \\
\midrule

MedVLThinker-7B~\cite{huang2025medvlthinker}
&55.00  & 54.12 & 27.50 & 57.50 & 63.50 & 64.00 & 53.60 \\
\midrule

OctoMed-7B~\cite{ossowski2025octomed}
& \underline{55.50}  & 56.47 & \textbf{35.00 }& 63.00 & \underline{84.00} & \textbf{79.00} & \underline{62.16} \\
\midrule

MedVLM-R1-2B~\cite{pan2025medvlm}
& 50.50 & 39.41 & 22.50 & 61.50 & 71.00 & 63.50 & 51.40 \\
\midrule

Med-R1-2B~\cite{lai2026med}
& 53.00 & 41.18 & 23.50 & \textbf{66.50} & 69.50 & 62.00 & 52.61 \\
\midrule

\rowcolor{winblue}
\textbf{Ours (7B)}
& \textbf{59.00} & \textbf{62.94} & \underline{34.50} & \underline{65.50} & \textbf{88.0} & \textbf{79.00 }& \textbf{64.91} \\

\bottomrule
\end{tabular}
}
\label{tab:med_vqa}
\end{table*}

\section{Conclusion}
We examined when RL post-training helps medical VLMs by separating the effects of vision, medical SFT, and RL. On MedMNIST, we observe a clear support boundary: Accuracy@1 can lag far behind Pass@K, indicating latent capability that greedy decoding does not realize. Medical SFT reliably improves both perception and support, while RL mainly sharpens the output distribution, improving Accuracy@1 when baseline support is already non-trivial and offering limited benefits under larger shifts.

These findings motivate a boundary-aware recipe: diagnose support with Pass@K, bridge weak-support regimes with targeted finetuning, then apply RL to improve sampling efficiency.

%
%
%
\clearpage
\bibliographystyle{splncs04}
\bibliography{main}

\begin{thebibliography}{10}
\providecommand{\url}[1]{\texttt{#1}}
\providecommand{\urlprefix}{URL }
\providecommand{\doi}[1]{https://doi.org/#1}

\bibitem{baghbanzadeh2025advancing}
Baghbanzadeh, N., Fallahpour, A., Parhizkar, Y., Ogidi, F., Roy, S., Ashkezari, S., Khazaie, V.R., Colacci, M., Etemad, A., Afkanpour, A., et~al.: Advancing medical representation learning through high-quality data. In: International Conference on Medical Image Computing and Computer-Assisted Intervention. pp. 24--33. Springer (2025)

\bibitem{bai2025qwen2}
Bai, S., Chen, K., Liu, X., Wang, J., Ge, W., Song, S., Dang, K., Wang, P., Wang, S., Tang, J., et~al.: {Qwen2.5-VL Technical Report}. arXiv preprint arXiv:2502.13923  (2025)

\bibitem{chen2024towards}
Chen, J., Gui, C., Ouyang, R., Gao, A., Chen, S., Chen, G.H., Wang, X., Cai, Z., Ji, K., Wan, X., et~al.: Towards injecting medical visual knowledge into multimodal llms at scale. In: Proceedings of the 2024 conference on empirical methods in natural language processing. pp. 7346--7370 (2024)

\bibitem{chen2025grpo}
Chen, Y., Ge, Y., Wang, R., Ge, Y., Cheng, J., Shan, Y., Liu, X.: Grpo-care: Consistency-aware reinforcement learning for multimodal reasoning. arXiv preprint arXiv:2506.16141  (2025)

\bibitem{dai2025qoq}
Dai, W., Chen, P., Ekbote, C., Liang, P.P.: Qoq-med: Building multimodal clinical foundation models with domain-aware grpo training. arXiv preprint arXiv:2506.00711  (2025)

\bibitem{fan2025chestx}
Fan, Z., Liang, C., Wu, C., Zhang, Y., Wang, Y., Xie, W.: Chestx-reasoner: Advancing radiology foundation models with reasoning through step-by-step verification. arXiv preprint arXiv:2504.20930  (2025)

\bibitem{he2020pathvqa}
He, X., Zhang, Y., Mou, L., Xing, E., Xie, P.: Pathvqa: 30000+ questions for medical visual question answering. arXiv preprint arXiv:2003.10286  (2020)

\bibitem{huang2025medvlthinker}
Huang, X., Wu, J., Liu, H., Tang, X., Zhou, Y.: Medvlthinker: Simple baselines for multimodal medical reasoning. arXiv preprint arXiv:2508.02669  (2025)

\bibitem{jeddi2025puzzle}
Jeddi, A., Karaimer, H.C., Nguyen, H., Wang, Z., Zhao, K., Rajabi, J., Zhang, R., Goyal, R., Taati, B., Grzeszczuk, R.: Puzzle curriculum grpo for vision-centric reasoning. arXiv preprint arXiv:2512.14944  (2025)

\bibitem{karan2025reasoning}
Karan, A., Du, Y.: Reasoning with sampling: Your base model is smarter than you think. arXiv preprint arXiv:2510.14901  (2025)

\bibitem{lai2026med}
Lai, Y., Zhong, J., Li, M., Zhao, S., Li, Y., Psounis, K., Yang, X.: Med-r1: Reinforcement learning for generalizable medical reasoning in vision-language models. IEEE Transactions on Medical Imaging  (2026)

\bibitem{lau2018dataset}
Lau, J.J., Gayen, S., Ben~Abacha, A., Demner-Fushman, D.: A dataset of clinically generated visual questions and answers about radiology images. Scientific data  \textbf{5}(1),  180251 (2018)

\bibitem{liu2021slake}
Liu, B., Zhan, L.M., Xu, L., Ma, L., Yang, Y., Wu, X.M.: Slake: A semantically-labeled knowledge-enhanced dataset for medical visual question answering. In: 2021 IEEE 18th international symposium on biomedical imaging (ISBI). pp. 1650--1654. IEEE (2021)

\bibitem{liu2025beyond}
Liu, C., Wang, H., Pan, J., Wan, Z., Dai, Y., Lin, F., Bai, W., Rueckert, D., Arcucci, R.: Beyond distillation: Pushing the limits of medical llm reasoning with minimalist rule-based rl. arXiv preprint arXiv:2505.17952  (2025)

\bibitem{manzari2025medical}
Manzari, O.N., Asgariandehkordi, H., Koleilat, T., Xiao, Y., Rivaz, H.: Medical image classification with kan-integrated transformers and dilated neighborhood attention. Applied Soft Computing p. 114045 (2025)

\bibitem{myronenko2025reasoning}
Myronenko, A., Yang, D., Turkbey, B., Aboian, M., Azamat, S., Akcicek, E., Yin, H., Molchanov, P., Edgar, M., He, Y., et~al.: Reasoning visual language model for chest x-ray analysis. arXiv preprint arXiv:2510.23968  (2025)

\bibitem{ossowski2025octomed}
Ossowski, T., Zhang, S., Liu, Q., Qin, G., Tan, R., Naumann, T., Hu, J., Poon, H.: Octomed: Data recipes for state-of-the-art multimodal medical reasoning. arXiv preprint arXiv:2511.23269  (2025)

\bibitem{pan2025medvlm}
Pan, J., Liu, C., Wu, J., Liu, F., Zhu, J., Li, H.B., Chen, C., Ouyang, C., Rueckert, D.: Medvlm-r1: Incentivizing medical reasoning capability of vision-language models (vlms) via reinforcement learning. In: International Conference on Medical Image Computing and Computer-Assisted Intervention. pp. 337--347. Springer (2025)

\bibitem{radford2021learning}
Radford, A., Kim, J.W., Hallacy, C., Ramesh, A., Goh, G., Agarwal, S., Sastry, G., Askell, A., Mishkin, P., Clark, J., et~al.: Learning transferable visual models from natural language supervision. In: International conference on machine learning. pp. 8748--8763. PmLR (2021)

\bibitem{wang2025medical}
Wang, W., Ma, Z., Ding, M., Zheng, S., Liu, S., Liu, J., Ji, J., Chen, W., Li, X., Shen, L., et~al.: Medical reasoning in the era of llms: a systematic review of enhancement techniques and applications. arXiv preprint arXiv:2508.00669  (2025)

\bibitem{xia2025mmedagent}
Xia, P., Wang, J., Peng, Y., Zeng, K., Wu, X., Tang, X., Zhu, H., Li, Y., Liu, S., Lu, Y., et~al.: Mmedagent-rl: Optimizing multi-agent collaboration for multimodal medical reasoning. arXiv preprint arXiv:2506.00555  (2025)

\bibitem{xu2025medground}
Xu, H., Nie, Y., Wang, H., Chen, Y., Li, W., Ning, J., Liu, L., Wang, H., Zhu, L., Liu, J., et~al.: Medground-r1: Advancing medical image grounding via spatial-semantic rewarded group relative policy optimization. In: International Conference on Medical Image Computing and Computer-Assisted Intervention. pp. 391--401. Springer (2025)

\bibitem{xu2025lingshu}
Xu, W., Chan, H.P., Li, L., Aljunied, M., Yuan, R., Wang, J., Xiao, C., Chen, G., Liu, C., Li, Z., et~al.: Lingshu: A generalist foundation model for unified multimodal medical understanding and reasoning. arXiv preprint arXiv:2506.07044  (2025)

\bibitem{medmnistv2}
Yang, J., Shi, R., Wei, D., Liu, Z., Zhao, L., Ke, B., Pfister, H., Ni, B.: Medmnist v2-a large-scale lightweight benchmark for 2d and 3d biomedical image classification. Scientific Data  \textbf{10}(1), ~41 (2023)

\bibitem{yue2024mmmu}
Yue, X., Ni, Y., Zhang, K., Zheng, T., Liu, R., Zhang, G., Stevens, S., Jiang, D., Ren, W., Sun, Y., et~al.: Mmmu: A massive multi-discipline multimodal understanding and reasoning benchmark for expert agi. In: Proceedings of the IEEE/CVF conference on computer vision and pattern recognition. pp. 9556--9567 (2024)

\bibitem{yue2025does}
Yue, Y., Chen, Z., Lu, R., Zhao, A., Wang, Z., Song, S., Huang, G.: Does reinforcement learning really incentivize reasoning capacity in llms beyond the base model? arXiv preprint arXiv:2504.13837  (2025)

\bibitem{zhang2025interplay}
Zhang, C., Neubig, G., Yue, X.: On the interplay of pre-training, mid-training, and rl on reasoning language models. arXiv preprint arXiv:2512.07783  (2025)

\bibitem{zhang2025med}
Zhang, S., Liu, Q., Qin, G., Naumann, T., Poon, H.: Med-rlvr: Emerging medical reasoning from a 3b base model via reinforcement learning. arXiv preprint arXiv:2502.19655  (2025)

\bibitem{zhang2025patho}
Zhang, W., Zhang, P., Guo, J., Cheng, T., Chen, J., Zhang, S., Zhang, Z., Yi, Y., Bu, H.: Patho-r1: A multimodal reinforcement learning-based pathology expert reasoner. arXiv preprint arXiv:2505.11404  (2025)

\bibitem{zhang2023pmc}
Zhang, X., Wu, C., Zhao, Z., Lin, W., Zhang, Y., Wang, Y., Xie, W.: Pmc-vqa: Visual instruction tuning for medical visual question answering. arXiv preprint arXiv:2305.10415  (2023)

\bibitem{zuo2025medxpertqa}
Zuo, Y., Qu, S., Li, Y., Chen, Z., Zhu, X., Hua, E., Zhang, K., Ding, N., Zhou, B.: Medxpertqa: Benchmarking expert-level medical reasoning and understanding. arXiv preprint arXiv:2501.18362  (2025)

\end{thebibliography}

\end{document}